\documentclass[letterpaper, 10 pt, journal, twoside]{URL-ieeeconf}

\usepackage{graphics}           
\usepackage{times}              
\usepackage{amsmath}            
\usepackage{amssymb}            
\usepackage{graphicx}
\usepackage{algorithm}
\usepackage[noend]{algpseudocode}
\usepackage{booktabs}
\usepackage{color}
\usepackage{multirow}
\usepackage{subcaption}
\usepackage{rotating}
\usepackage{cite}

\usepackage{hyperref}
\usepackage{dsfont}
\usepackage{siunitx}    
\definecolor{instructioncolor}{rgb}{.5,.5,.5}
\definecolor{addcolor}{rgb}{0,0,1}

\usepackage[font=small]{caption}

\def\secref#1{Section~\ref{#1}}
\def\figref#1{Fig.~\ref{#1}}
\def\tabref#1{Table~\ref{#1}}
\def\eqref#1{(\ref{#1})}

\newcommand{\rom}[1]{\uppercase\expandafter{\romannumeral #1\relax}}

\makeatletter
\usepackage{xspace}
\DeclareRobustCommand\onedot{\futurelet\@let@token\@onedot}
\def\@onedot{\ifx\@let@token.\else.\null\fi\xspace}

\makeatother

\usepackage{array}
\newcolumntype{L}[1]{>{\raggedright\let\newline\\\arraybackslash\hspace{0pt}}m{#1}}
\newcolumntype{C}[1]{>{\centering\let\newline\\\arraybackslash\hspace{0pt}}m{#1}}
\newcolumntype{R}[1]{>{\raggedleft\let\newline\\\arraybackslash\hspace{0pt}}m{#1}}

\newcommand{\norm}[1]{\lVert#1\lVert}

\newcommand{\degrees}{{\mbox{$^\circ$}}}

\newcommand{\envQ}{\mathbf{\mathcal{Q}}}
\newcommand{\varGraph}{\mathbf{G}}
\newcommand{\varGraphNode}{\mathbf{V}}
\newcommand{\varGraphEdge}{\mathbf{E}}
\def\varNode#1{\mathbf{v}_{#1}}                  
\def\varPosVector#1{\mathbf{p}_{#1}}    
\def\varState#1{s_{#1}}                 
\def\varEdgeSet#1{E_{#1}}     
\newcommand{\varCircle}{c_i}  
\newcommand{\varMapResolution}{\rho_\text{map}} 
\newcommand{\varRobotsize}{r_{\text{robot}}}      
\newcommand{\stValid}{\text{valid}}     
\newcommand{\stInvalid}{\text{invalid}} 
\newcommand{\stFrontier}{\text{frontier}}   
\newcommand{\fnStability}{g(\varCircle)}    
\newcommand{\fnReachability}{r(\varEdgeSet{i})} 
\def\fnNumberEdges#1{n(#1)} 
\def\fnHeight#1{h(x_{#1},y_{#1})}                
\def\varMedianHeight{h_{\text{mid}}}                     
\def\varMaxHeight{h_{\text{max}}}              

\def\varEdge#1#2{e_{#1#2}}           
\def\varDist#1#2{d_{#1#2}}                  
\def\varWeight#1#2{w_{#1#2}}                
\newcommand{\varEllipse}{\xi}           
\newcommand{\varRiskRatio}{\gamma}
\def\varRisk#1{\mathcal{R}^\varEllipse_{\text{#1}}}
\def\varEigen#1{\mathbf{\hat{e}^\varEllipse_{\text{#1}}}}
\def\varPathDir#1{\mathbf{\hat{n}}_{\text{#1}}}
\newcommand{\varGround}{\mathbf{g}}

\newcommand{\varExpRadius}{r_{\text{exp}}}

\def\varSubNode#1{V_{#1}}
\def\varOutVector#1{\mathbf{u}_{#1}}

\def\fnO#1{C (#1)}
\def\fnJ#1{J (#1)}
\newcommand{\safetyGain}{\Gamma}

\def\lowerRoman#1{\lowercase{\expandafter{\romannumeral#1\relax}}}
\def\upperRoman#1{\uppercase{\expandafter{\romannumeral#1\relax}}}

\newcommand{\meterUnit}{\mathrm{m}}
\newcommand{\percentUnit}{\mathrm{\%}}
\newcommand{\secUnit}{\mathrm{s}}
\newcommand{\msecUnit}{\mathrm{ms}}

\newcommand{\srp}{\mathcal{S}_{\text{path}}}
\newcommand{\srt}{\mathcal{S}_{\text{trav}}}

\newcommand{\pathLength}{\mathcal{L}_{\text{path}}}
\newcommand{\travLength}{\mathcal{L}_{\text{trav}}}
\newcommand{\travPathDeviation}{\mathcal{T}}
\newcommand{\pathRisk}{\mathcal{W}}

\def\tbf#1{\textbf{#1}}
\def\uli#1{\underline{#1}}
\def\mc#1{\multicolumn{1}{l|}{#1}}

\newcommand{\Seshort}{\texttt{Short} }
\newcommand{\Semedium}{\texttt{Medium} }
\newcommand{\Selong}{\texttt{Long} }

\title{\LARGE \bf TRG-planner: Traversal Risk Graph-Based Path Planning in Unstructured Environments for Safe and Efficient Navigation}

\author{Dongkyu Lee, I Made Aswin Nahrendra, Minho Oh, Byeongho Yu, and Hyun Myung$^{*}$
  \thanks{$^*$Corresponding author: Hyun Myung}
  \thanks{The authors are with the School of Electrical Engineering, KAIST (Korea Advanced Institute of Science and Technology), Daejeon, 34141, Republic of Korea. {\tt\scriptsize \{dklee, anahrendra, minho.oh, bhyu, hmyung\}@kaist.ac.kr} \hfill \break   	 
  \indent 
  }
}

\begin{document}
\maketitle
\thispagestyle{empty}
\pagestyle{empty}

\begin{abstract}
  Unstructured environments such as mountains, caves, construction sites, or disaster areas are challenging for autonomous navigation because of terrain irregularities.
  In particular, it is crucial to plan a path to avoid risky terrain and reach the goal quickly and safely.
  In this paper, we propose a method for safe and distance-efficient path planning, leveraging Traversal Risk Graph~(TRG), a novel graph representation that takes into account geometric traversability of the terrain.
  TRG nodes represent stability and reachability of the terrain, while edges represent relative traversal risk-weighted path candidates. 
  Additionally, TRG is constructed in a wavefront propagation manner and managed hierarchically, enabling real-time planning even in large-scale environments.
  Lastly, we formulate a graph optimization problem on TRG that leads the robot to navigate by prioritizing both safe and short paths. 
  Our approach demonstrated superior safety, distance efficiency, and fast processing time compared to the conventional methods.
  It was also validated in several real-world experiments using a quadrupedal robot. 
  Notably, TRG-planner contributed as the global path planner of an autonomous navigation framework for the DreamSTEP team, which won the Quadruped Robot Challenge at ICRA 2023. The project page is available at \url{https://trg-planner.github.io}.
\end{abstract}



\section{Introduction}
\label{sec:intro}

\PARstart{S}{afe} and efficient path planning is an essential component for mobile robots in successful autonomous navigation.
Traditionally, a safe path is defined as a collision-free path from the start to the goal position~\cite{yang2022far, kim2022learning, jian2021global}.
However, the concept of a safe path in unstructured environments such as mountains, caves, construction sites, or disaster areas becomes ambiguous due to irregular terrain.
That is because the safety of the path is affected not only by obstacle collisions but also due to terrain properties such as steepness and roughness.

Thanks to recent advancements in mobile robot platforms~\cite{nakajima2011rt, xu2020high} or in the locomotion ability of legged robots~\cite{nahrendra2023dreamwaq, kareer2023vinl, choi2023learning}, which can overcome rough terrain, robots can navigate such unstructured environments.
However, even with robust locomotion ability, they can inevitably fail to navigate due to the unsafe planned path on rough terrain.
Therefore, in planning the navigation path, the traversal risk of the terrain should be considered to prevent unstable navigation that yields navigation failure.

To avoid the unstable terrains, some researchers proposed the traversability mapping methods to identify the safe area~\cite{frey2022locomotion, miki2022elevation, shan2018bayesian}.
The existing methods usually represent traversability based on the terrain stability, which means the robot can stand still on the terrain.
However, the primary purpose of these methods is to improve the map quality during navigation with partial sensor measurements.
Therefore, these methods are insufficient to represent the relative risk of path candidates, which can be different depending on the entry direction into the terrain.
For example, the robot can maintain stability when entering a slope from the front, but it may become unstable when entering from the side.
Additionally, reachability, which refers to the possibility that the robot can reach the goal location from its current pose, is not considered in the traversability map and should be checked during navigation.

\begin{figure}[t!]
    \captionsetup{font=footnotesize}
	\centering
    \includegraphics[width=\columnwidth]{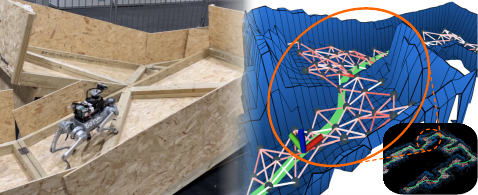}
    \caption{
        (L-R): Our quadruped robot autonomously navigates through a harsh slope terrain during the QRC competition. 
        The proposed planner constructs the traversal risk graph, which contains the reachability of the terrain and the relative risk of the path candidates. 
        The white-to-red gradation of the graph edges visualizes the relative risk of the path candidates. 
        The robot plans a safe and efficient path by optimizing the graph (green line).
        }
    \label{fig:01_main_qrc}
\end{figure}
\begin{figure*}[t!]
    \captionsetup{font=footnotesize}
	\centering
    \includegraphics[width=0.9\textwidth]{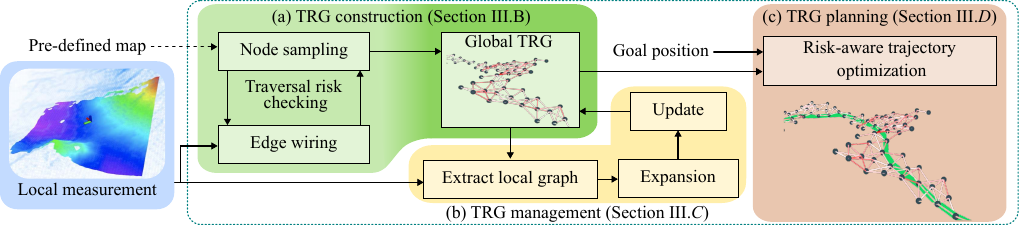}
    \caption{
        Overview of our proposed path planner framework called \textit{TRG-planner}.
        Traversal Risk Graph (TRG) is proposed to capture the geometrical information of the terrain, consisting of nodes, edges, and their relationships (\secref{sec:TRG_components}).
        (a) TRG is constructed by sampling nodes from the local elevation map or, optionally, from a prebuilt map (dashed line) and connecting edges based on the relative risk of path candidates.
        (b) The local graph is extracted from the global graph. It is hierarchically expanded and updated.
        (c) TRG-planner optimizes the risk-aware cost function to find a safe and distance-efficient path.
    }
    \label{fig:framework}
    \vspace{-0.3cm}
\end{figure*}

In this context, we propose a planning method to find a safe and efficient path by introducing a novel graph structure called Traversal Risk Graph (TRG) that models the geometrical information of the terrain, as shown in Fig.~\ref{fig:01_main_qrc}.
TRG consists of nodes and edges. 
The nodes represent safe areas, determined by the stability of the corresponding terrain and the reachability from the current robot pose. 
The edges indicate traversal risk-weighted path candidates between the nodes.
The proposed graph sequentially propagates nodes and wires edges in local area and combines them into a global structure, enabling time-efficient hierarchical management of all reachable environments.
We formulate the safe path planning problem as a graph optimization problem on TRG, prioritizing paths with low risk and short distances.
Our proposed planning method, TRG-planner, outperforms conventional methods in terms of safety and distance-efficiency in simulation environments.
Furthermore, TRG-planner has been validated in real-world environments using a quadrupedal robot.
In summary, the main contributions of this paper are as follows:

\begin{itemize}
    \item 
        Our proposed Traversal Risk Graph (TRG) configures the traversable environments effectively, representing the relative risk and reachability of path candidates in real-time.
    \item 
        TRG-planner introduces novel terrain sampling and graph construction strategies that efficiently represent the traversal properties of unstructured environments.
    \item
        TRG-planner generates a safe path by optimizing a risk-aware cost function and has been successfully validated in real-world unstructured environments, including the Quadruped Robot Challenge (QRC) at ICRA 2023.
\end{itemize}

\section{Related Works}
\label{sec:related_works}

\subsection{Map Representation For Path Planning}
Environment representation is one of the key issues for map-based path planning.
Various existing map representations~\cite{elfes1989computer, hornung2013ar, krusi2017driving, oleynikova2017voxblox, fankhauser2014robot, fankhauser2018probabilistic} are selected according to the environments, sensors, and the purpose of the navigation.
2D occupancy grid maps~\cite{elfes1989computer} are widely used because they simply classify the environment into occupied space which should be avoided by the robot, and free space which can be traversed by the robot.
Although this binary classification is suitable for indoor environments where the robot moves on flat ground and treats anything with height as an obstacle, it lacks information for complex environments with non-flat terrains.

In contrast, 3D map representations~\cite{hornung2013ar, krusi2017driving, oleynikova2017voxblox} can represent full 3D environments.
However, additional computational resources to extract the traversable ground~\cite{oh2022travel} or configuration space~\cite{brandao2020gaitmesh} are required for mobile robots that navigate only on the planar surfaces.


As a middle ground between 2D and 3D, the 2.5D elevation map~\cite{fankhauser2014robot, fankhauser2018probabilistic} was proposed to represent the heights of the environments.
Owing to its capability to represent the characteristics of the terrain, an elevation map is often selected to represent unstructured environments, including rough terrain, slopes, and bumps~\cite{wermelinger2016navigation}.
In this study, we focus on unstructured environments such as mountainous and rough terrains, and thus, represent the environment using the elevation maps.

\subsection{Traversablility-Aware Path Planning}
Compared to navigation on flat terrain, unstructured terrain is ambiguous as it is difficult to determine whether it is a collision space or not. 
To address this problem, several approaches have been proposed to estimate traversability from diverse aspects.
Geometrical approaches consider the geometrical properties of the terrain such as the slope, roughness, or bumpiness~\cite{wermelinger2016navigation,norby2020fast, fan2021step, chen2023smug, liu2023hybrid, yoo2024traversability}.
Geometrical traversability maps can be utilized for path planning because they have safe terrain information.
However, they often estimate the traversability in one direction on a terrain without any consideration on terrain entry direction nor reachability.

Several approaches have attempted to process the semantic information of the terrain that cannot be known through geometric information~\cite{wellhausen2019should, wellhausen2020safe, guan2022ga}.
Although effective for diverse terrains like mud, water, and grass, these approaches require extensive labeled data and limit the generalization to unseen data.
Recently, learning-based methods have also been used to estimate the motion of the robot to predict the traversability~\cite{guzzi2020path, wellhausen2021rough, wellhausen2023artplanner, yang2021real}.
These methods aim to locally optimize the robot motion considering a given global path.
However, if a given global path is not safe, the robot could still be at risk even if it can locally optimize the path safely.

Our work focuses on the geometrical traversability through the proposed graph structure, aiming to plan a globally safe and short path in unstructured environments.
Therefore, the proposed method can serve as the global path planner with simple path followers and be combined with the aforementioned semantic or local optimization methods in the autonomous navigation framework.




\section{Traversal Risk Graph}\label{sec:Method}

The main objective of finding a safe path is to minimize the risk of the robot falling, misstepping, or staggering.
We address this problem by proposing a Traversal Risk Graph (TRG), and its framework is described in \figref{fig:framework}.
TRG consists of nodes and edges that effectively represent the traversal properties (\secref{sec:TRG_components}).
TRG is initialized using a sporadic spreadable sampling method (\secref{sec:TRG_construction}).
Then, TRG is managed hierarchically (\secref{sec:TRG_management}).
Finally, the safe path is optimized by the risk-aware cost function, leveraging TRG (\secref{sec:TRG_planning}).

\subsection{TRG Components}\label{sec:TRG_components}

The environment, denoted as $\envQ \in \mathbb{R}^3$, serves as the workspace for robot navigation.
TRG is designed to represent $\envQ$ in a manner that is suitable for path planning, considering the relative risk and reachability of the path candidates, rather than handling the geometric details of the entire environment.
TRG interprets the 3D space $\envQ$ as a lower-dimensional undirected graph structure $\varGraph = (\varGraphNode, \varGraphEdge)$ consisting of a set of nodes $\varGraphNode$ and edges $\varGraphEdge$.

\subsubsection{Node}
Each node~$\varNode{i} = (\varPosVector{i}, \varState{i}, \varEdgeSet{i}) \in \varGraphNode$ represents an area that the robot can stand on and reach from its current location. Here, $\varPosVector{i} \in \mathbb{R}^3$ is the position of the node, $\varState{i}$ is the state of the node, and $\varEdgeSet{i}$ is the subset of edges~$\varGraphEdge$ connected to the node~$\varNode{i}$.
A corresponding terrain region is defined as an inscribed circle~$\varCircle$ that fits within the robot body with radius of~$\varRobotsize$, with $\varPosVector{i}$ being the center of $\varCircle$.
To decide the state of the node~$\varState{i}$, the robot's absolute geometrical stability~$\fnStability$ and node reachability~$\fnReachability$ metrics are defined as follows:
\begin{equation}\label{eq:stability_reachability}
    \begin{aligned}
    \fnStability & = \mathds{1}{\{\forall \fnHeight{k} : \; |\fnHeight{k} - \varMedianHeight| < \varMaxHeight\}}, \\
    \fnReachability & = \mathds{1}{\{\fnNumberEdges{\varEdgeSet{i}} > 0\}}, \\
    \end{aligned}
\end{equation}
where $\fnHeight{k} = \{z_{k} \, | \, (x_{k},y_{k}) \in \varCircle\}$, $\varMedianHeight$ is the median height of the area~$\varCircle$, $\varMaxHeight$ is the maximum height threshold to overcome, and $\fnNumberEdges{\varEdgeSet{i}}$ is the number of edges connected to the node~$\varNode{i}$, respectively.
Both $\fnStability$ and $\fnReachability$ are formulated as binary indicator functions~$\mathds{1}{\{\text{condition}\}}$ that return 1 if the condition is satisfied, and 0 otherwise.
Therefore, $\fnStability$ determines whether the node area is collision-free or too rough to stand on, and $\fnReachability$ is used to check whether the robot can reach the node~$\varNode{i}$ by passing connected edges.
Consequently, the state of the node~$\varState{i}$ is classified as \stValid \space or \stInvalid \space by combining the stability and reachability metrics as follows:
\begin{equation}\label{eq:node_state}
    \varState{i} =
    \begin{cases}
        \stValid, & \text{if } \fnStability \wedge \fnReachability , \\
        \stInvalid, & \text{otherwise}.
    \end{cases}
\end{equation}

\subsubsection{Edge}
The edge~$\varEdge{i}{j} = (\varNode{j}, \varDist{i}{j}, \varWeight{i}{j}) \in \varEdgeSet{i}$ represents the path candidates that can travel from the node $\varNode{i}$ to the node $\varNode{j}$ which is the destination node, where $\varDist{i}{j} = \norm{\varPosVector{i} - \varPosVector{j}}$ and $\varWeight{i}{j}$ is the weight of the edge. 
Notably, $\varWeight{i}{j}$ represents the potential relative risk for the robot to lose stability or encounter obstacles that could hinder its path.

\begin{figure}[t!]
    \captionsetup{font=footnotesize}
	\centering
    \includegraphics[width=\columnwidth]{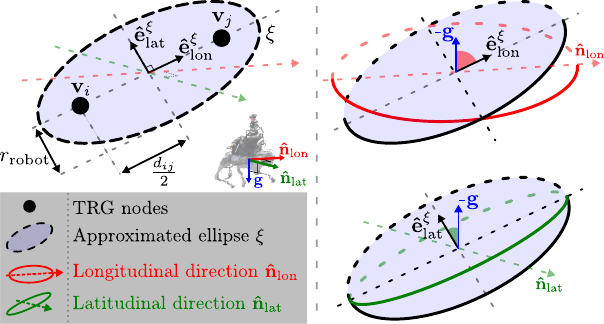}
    \caption{
        The edge area is approximated to an ellipse plane~$\varEllipse$ using principal component analysis (PCA).
        Nodes~$\varNode{i}$ and~$\varNode{j}$ are the focal points of the ellipse, and the minor axis of the ellipse is determined by the radius~$\varRobotsize$ of the inscribed circle of the robot.
        $\varPathDir{lon}$ and $\varPathDir{lat}$ are the longitudinal and latitudinal unit direction vectors along the path, and $\varEigen{lon}$ and~$\varEigen{lat}$ are normalized eigenvectors of the ellipse along each direction, respectively.
    }
    \label{fig:edge_weight}
\end{figure}

To quantify $\varWeight{i}{j}$, the region near the path is approximated to an ellipse plane~$\varEllipse$ using principal component analysis (PCA) of the measured height map, as shown in \figref{fig:edge_weight}.
First, the edge~$\varEdge{i}{j}$ can be wired only if three conditions are satisfied: (\lowerRoman{1}) the region forming $\varEllipse$ has more than three height points, (\lowerRoman{2}) all deviations of the height points in $\varEllipse$ are less than $\varMaxHeight$, (\lowerRoman{3}) the edge is not too steep, i.e. satisfies the following equation:
\begin{equation}\label{eq:edge_condition}
    \tan^{-1}{\left(\frac{|\varPosVector{i}(z) - \varPosVector{j}(z)|}{\norm{\varPosVector{i}(x,y) - \varPosVector{j}(x,y)}}\right)} < \tan^{-1}{\left(\frac{\varMaxHeight}{\varRobotsize}\right)}.
\end{equation}
Then, $\varWeight{i}{j}$ of a valid edge is defined as:
\begin{equation}
    \label{eq:edge_weight}
    \begin{aligned}
        \varWeight{i}{j} & = \varRiskRatio \varRisk{lon} + (1 - \varRiskRatio) \varRisk{lat}, \\
        \varRisk{dir} & = -\varEigen{dir} \cdot \varGround, \; \text{dir} \in \{\text{lon}, \text{lat}\},
    \end{aligned}
\end{equation}
where $\varRisk{lon}$ and $\varRisk{lat}$ represent the risk along the longitudinal and latitudinal directions along the path candidate, respectively. 
Here, $\varEigen{lon}$ and $\varEigen{lat}$ denote the normalized eigenvectors of the ellipse plane, and $\varRiskRatio$ is the ratio of the risk along the longitudinal direction to the total risk.
$\varRisk{lon}$ and $\varRisk{lat}$ are calculated by the negative inner product of the eigenvectors with the gravity vector~$\varGround = [0, 0, -1]$.
In other words, risk~$\varWeight{i}{j}$ reflects the larger likelihood of a falling robot when the path is inclined to the corresponding direction.

Consequently, regions represented by nodes can be considered either reachable or unreachable depending on the edge selection, as the relative risk~$\varWeight{i}{j}$ determines the traversal difficulty based on the entry direction into the terrain.
For example, it may be challenging for the robot to reach node~$\varNode{j}$ through an edge in the forward direction with a high-risk weight.
However, a combination of alternative directional edges with lower weights ($\varEdge{i}{*}$ and $\varEdge{*}{j}$) may enable reaching~$\varNode{j}$ safely.
In such scenarios, $\varNode{j}$ is considered reachable, subject to the constraint of selecting safe directional edges.

\subsection{TRG Construction}\label{sec:TRG_construction}
TRG is constructed in a sampling-based manner, similar to the probabilistic roadmap (PRM)~\cite{kavraki1996probabilistic}.
However, rather than sampling on the entire environment~$\envQ$, TRG is initialized by sampling nodes in the vicinity of the reference node~$\varNode{\text{ref}}$, which indicates the current robot position.
In other words, nodes are added incrementally outward from the reference node, similar to wavefront propagation.
This strategy aims to include only the regions into TRG that are navigable from the robot’s current position, significantly enhancing the time efficiency of graph construction.

\begin{figure}[t!]
    \captionsetup{font=footnotesize}
    \centering
    \includegraphics[width=0.95\columnwidth]{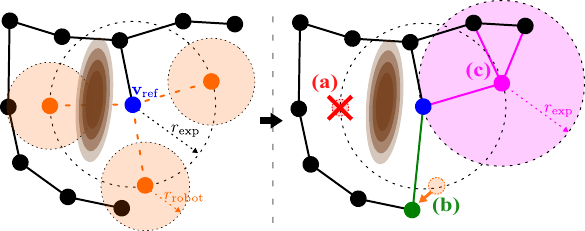}
    \caption{
        Example of TRG expansion. 
        (Left) Nodes (orange dots) are randomly sampled from a uniform distribution on a circle with a radius of~$\varExpRadius$ from the reference node~$\varNode{\text{ref}}$ (blue dot).
        (Right) There are three possible cases for the sampled nodes.
        (a) Sampled nodes are discarded if the corresponding terrain is unstable or if the terrain between the reference node and the sampled node is unsuitable for traversal.
        (b) If an existing node is within the radius of~$\varRobotsize$ from the sampled node, they are merged and an edge is wired (green line) between the reference node and the existing node (green dot).
        (c) Otherwise, the sampled node is generated (magenta dot), and edges are subsequently wired between the sampled and existing node within~$\varExpRadius$ (magenta lines).
        }
    \label{fig:graph_expansion}
    \vspace{-0.2cm}
\end{figure}

The graph construction process is divided into three steps: sampling nodes, checking traversal risk, and wiring edges.
Initially, nodes are randomly sampled following a uniform distribution on a circle with a radius of the node expansion radius~$\varExpRadius$, centered at~$\varPosVector{\text{ref}}$.
This is because paths that are too short are unsuitable for approximating the plane to estimate $\varWeight{i}{j}$, and overly long paths cannot accurately represent the geometrical terrain property.
Once the node is sampled, the traversal risk is checked by \eqref{eq:node_state} and \eqref{eq:edge_condition}.
As depicted in \figref{fig:graph_expansion}, there are three possible cases for the sampled node:
(\lowerRoman{1}) The sampled node is discarded if it is determined to be in an \stInvalid \space state by \eqref{eq:node_state}.
(\lowerRoman{2}) If an existing node is within the radius of~$\varRobotsize$ from the sampled node, it is merged with the existing node, and an edge is wired between the $\varNode{\text{ref}}$ and the existing node.
(\lowerRoman{3}) Otherwise, the sampled node is added to both the nodes~$\varGraphNode$ and the expansion queue, with an edge subsequently wired between the $\varNode{\text{ref}}$ and the sampled node.
Additionally, all nodes within~$\varExpRadius$ from the sampled node are also wired with the sampled node.
Finally, the front node of the expansion queue becomes a new reference node, and the process iterates until the expansion queue is empty, enabling an effective representation of all traversable space.

\subsection{TRG Management}\label{sec:TRG_management}
\begin{figure}[t!]
    \captionsetup{font=footnotesize}
    \centering
    \includegraphics[width=\columnwidth]{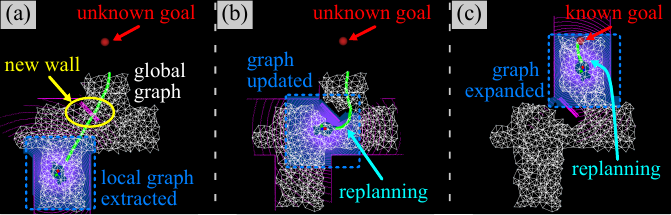}
    \caption{
        Example of hierarchical graph management.
        (a) The robot navigates along the planned path (green) to the sub-goal closest to the unknown goal (red). The local graph, shown in the blue dashed box, is extracted from the global graph which is shown in white. The invisible wall, highlighted as a yellow ellipsoid, is outside the local measurement.
        (b) Some nodes are updated to the \stInvalid \space states as the unseen wall is detected.
        Then, replanning is conducted because the previous path is on the \stInvalid \space node.
        (c) When the robot moves to the unknown area, the local graph is expanded from the \stFrontier \space nodes, leading to the global graph update.
        Finally, the sub-goal becomes the known goal position, and the robot reaches the goal.
        }
    \label{fig:graph_update}
    \vspace{-0.2cm}
\end{figure}

After TRG is initialized, the graph is updated hierarchically, as shown in~\figref{fig:graph_update}.
Basically, TRG is managed as a global graph, but the local nodes~$\varSubNode{l} \subset \varGraphNode$ are extracted in the vicinity of the robot from the local measurements.
During the update phase, every node in $\varSubNode{l}$ is also checked whether its state is \stFrontier \space as follows:
\begin{equation}
    \label{eq:frontier}
    \begin{aligned}
        \varSubNode{f} &= \bigcup_{\varNode{i} \in \varSubNode{l}} \bigl\{\varNode{i}  \mid \varState{i} = \stValid \wedge\varOutVector{i} \notin \envQ_\varGraph \bigr\}, \\
        \varOutVector{i} &= \varPosVector{i} + 2 \varRobotsize \frac{\varPosVector{i} - \varPosVector{\text{cur}}}{\norm{\varPosVector{i} - \varPosVector{\text{cur}}}}. \\
    \end{aligned}
\end{equation}
Here, $\varSubNode{f} \subset \varSubNode{l}$ is the set of \stFrontier \space nodes, $\envQ_\varGraph \subset \envQ$ is the area covered by~$\varGraph$, $\varPosVector{\text{cur}}$ is the current robot position, and $\varOutVector{i}$ is the vector from the \stValid \space nodes~$\varNode{i}$ in the direction away from the robot.
Therefore, \stFrontier \space nodes are determined simply by checking whether $\varOutVector{i}$ is out of~$\envQ_\varGraph$, which indicates that $\varNode{i}$ is a leaf node.
$\varSubNode{l}$ without \stFrontier \space nodes are updated by checking the stability and reachability of the nodes as in~\eqref{eq:node_state}.
The edges are also updated through~\eqref{eq:edge_condition} and~\eqref{eq:edge_weight} only between the valid local nodes.
Then, the graph is sporadically expanded using $\varSubNode{f}$ as reference nodes.
Finally, the updated and expanded $\varSubNode{l}$ is integrated into the global graph~$\varGraph$ = ($\varGraphNode$, $\varGraphEdge$) as:
\begin{equation}
    \label{eq:graph_update}
    \begin{aligned}
        \varGraphNode & \, \leftarrow \, \varGraphNode \cup \varSubNode{l}, \\
        \varGraphEdge & \, \leftarrow \, \varGraphEdge \cup \bigl\{\varEdge{i}{j} \mid \varNode{i}, \varNode{j} \in \varSubNode{l} \bigr\}.
    \end{aligned}
\end{equation}

\vspace{-0.2cm}
\subsection{TRG Planning}\label{sec:TRG_planning}
In the planning phase, any graph search algorithm such as A*~\cite{hart1968formal}, can be applied to find the path from the current robot position to the goal position.
Although TRG contains feasible paths that enable the robot to navigate safely, existing cost functions are insufficient for safe navigation as they typically optimize the distance only.
Therefore, we expand the cost function of A* to consider the potential traversal risk of the path as follows:
\begin{equation}\label{eq:TRG_cost_function}
    \begin{aligned}
        \fnO{\varNode{i+1}} &= \fnO{\varNode{i}} + \varDist{i+1,}{i}\,(\safetyGain \varWeight{i+1,}{i} + 1), \\
        \fnJ{\varNode{i+1}} &= \fnO{\varNode{i+1}} + \norm{\varPosVector{i+1} - \varPosVector{\text{goal}}},
    \end{aligned}
\end{equation}
where $\safetyGain$ is the safety factor that can adjust the safety level of the path, and $\varPosVector{\text{goal}}$ is the goal position.
$\fnO{\varNode{i+1}}$ is the cost of the path from the start to the searching goal node, consisting of the previous cost~$\fnO{\varNode{i}}$, the relative risk~$\varWeight{i+1,}{i}$ of the edge between $\varNode{i}$ and $\varNode{i+1}$ as calculated in~\eqref{eq:edge_weight}, and the Euclidean distance~$\varDist{i+1,}{i}$ between nodes.
Owing to the second term of $\fnO{\varNode{i+1}}$, the path can be obtained that minimizes both the distance and the risk of the path, with a larger $\safetyGain$ resulting in a safer path.
Therefore, $\fnJ{\varNode{i+1}}$ is the total cost of the path with the heuristic function that calculates the minimum distance to the goal position~$\varPosVector{\text{goal}}$.

When the goal position~$\varPosVector{\text{goal}}$ is queried, the path is determined by minimizing the cost function~$\fnJ{\varNode{i+1}}$.
However, if the goal position is not on TRG, the robot sets a sub-goal among the \stFrontier \space nodes that are closest to the goal position (see~\figref{fig:graph_update}).
Then, the sub-goal node becomes the goal node by repeating update process and replanning.
At the same time, if the planned path is invalid as TRG is updated, replanning is performed.

\begin{figure}[t!]
    \captionsetup{font=footnotesize}
    \centering
    \includegraphics[width=0.95\columnwidth]{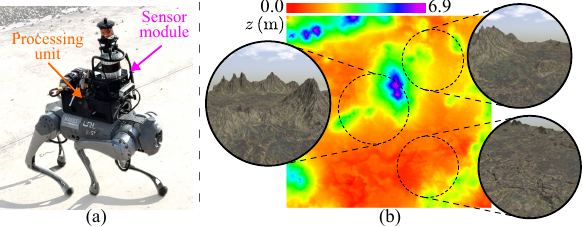}
    \caption{
        (a) Our quadruped robot, utilized in real-world environments, is equipped with a processing unit and a sensor module on top of the main body.
        (b) Examples of the simulation environment and the height map used in quantitative comparison, from a bird's-eye view.
        }
    \label{fig:simulation_environment}
    \vspace{-0.2cm}
\end{figure}
\vspace{-0.2cm}
\section{Experiments}\label{sec:experiments}
\vspace{-0.1cm}

We conducted experiments to demonstrate the effectiveness of the proposed TRG-planner, utilizing a Unitree Go1 quadruped robot, as shown in~\figref{fig:simulation_environment}(a), with a robust locomotion controller, DreamWaQ~\cite{nahrendra2023dreamwaq} in both the simulation and real-world environments.
The robot pose and local elevation map were obtained using slightly modified versions from~\cite{xu2022fast} and~\cite{fankhauser2018probabilistic}, respectively.
Once TRG-planner generated the path, the robot followed the path autonomously using the pure pursuit algorithm~\cite{coulter1992implementation}.

The parameters of the proposed method used in experiments are listed in~\tabref{tab:parameters}.
The resolution of the predefined map $\varMapResolution$ was unified for all experiments.
$\varRobotsize$, $\varExpRadius$, and $\varMaxHeight$ were set to robot-specific values: the width of the robot, the length of the robot, and the maximum height that can be overcome, respectively.
$\varRiskRatio$ is bounded between 0 and 1. 
$\safetyGain \in \mathbb{R}$ is a tunable parameter to adjust the balance between the distance and safety, with a lower bound of 0.
In this study, $\varRiskRatio$ and $\safetyGain$ were empirically set, but they can be adjusted according to the safety strategy.

\subsection{Experimental Setup} \label{sec:experimental_setup}
\subsubsection{Simulated Environment}
The simulated environment was a 50$\meterUnit$$\times$50$\meterUnit$$\times$6.9$\meterUnit$ wild mountainous area with irregular slopes (\figref{fig:simulation_environment}(b)).
Because the environment contained gentle and steep slopes, the robot must determine whether to pass through or detour around the terrain for safe and distance-efficient navigation.
We compared the proposed method with vanilla A*~\cite{hart1968formal}, PRM*~\cite{karaman2011sampling}, and T-Hybrid~\cite{liu2023hybrid} algorithms to demonstrate the improvement in safety and efficiency.
Furthermore, we compared the balance strategy ($\safetyGain=3.0$), the optimistic strategy ($\safetyGain=1.0$) and the conservative strategy ($\safetyGain=10.0$) of TRG-planner to investigate the balance between distance and safety according to the safety factor~$\safetyGain$.
To avoid non-smooth paths, we applied post-smoothing using moving average method~\cite{box2015time} to all compared methods.
Comparison was conducted at 2Hz on a desktop PC with an Intel Core i7-11700K CPU.

We set three scenarios based on the straight-line distance between the start and goal points: \Seshort (10$\meterUnit$), \Semedium (20$\meterUnit$), and \Selong (30$\meterUnit$) distances.
In each scenario, 100 start and goal positions and heading angles were randomly generated, and resampled whenever any generated positions were unsuitable for robot standing.

\subsubsection{Real-world Environments}
We tested our TRG-planner in three real-world environments: a mountainous environment, a mound environment, and an extremely difficult QRC arena~\cite{jacoff2023taking}.
First, the mountainous environment was a 83$\meterUnit$$\times$151$\meterUnit$$\times$38.3$\meterUnit$ area with various terrains, such as slopes, stairs, and narrow walkways, making the robot hard to decide on the appropriate path.
This environment was selected to test the long global path planning capability on unstructured terrains.
Second, the mound environment was 10$\meterUnit$$\times$14$\meterUnit$$\times$17.5$\meterUnit$ with different slopes according to the direction, only one of which was possible to climb.
It was designed to test the robot's ability to find the safe direction to climb the mound.
Finally, the QRC arena was tested, consisting of 5 sections: ramps, a soft floor with step-overs, steps with pipes, a floor with K-rails, and crate-mimic terrain.
Some sections even had slopes as steep as 15$\degrees$.
The robot should choose the safest path in each section to successfully navigate the arena.

\begin{table}[t!]
    \caption{
        Parameters of TRG-planner. 
        The units of $\varMapResolution$, $\varRobotsize$, $\varExpRadius$, and $\varMaxHeight$ are $[\meterUnit]$, and $\varRiskRatio$ and $\safetyGain$ are dimensionless.}
    \label{tab:parameters}
    \captionsetup{font=footnotesize}
    \centering
    \begin{tabular}{c|c|c|c|c|c|c}
        \toprule
        Parameter & $\varMapResolution$ & $\varRobotsize$ & $\varExpRadius$ & $\varMaxHeight$ & $\varRiskRatio$ & $\safetyGain$\\
        \midrule
        Value & 0.05 & 0.3 & 0.6 & 0.16 & 0.2 & 3.0 \\
        \bottomrule
    \end{tabular}
    \vspace{-0.2cm}
\end{table}

\subsection{Evaluation Metrics}
To evaluate planning method in terms of performance, safety, traveled distance, and time efficiency, we measured the path planning success rate ($\srp$), the travel success rate ($\srt$), the planned path length ($\pathLength$), and the planning time in the simulation environment.
Path planning success reflects the performance in planning over a variety of terrains, such as slopes and narrow sidewalks. 
Travel success, defined as reaching the goal location without the robot's body touching the ground, indicates how safe the planned path is for the robot to travel stably.
Additionally, we introduced two new metrics to measure the safety of the path when the robot successfully travels.
The traveled path deviation ($\travPathDeviation$) and the normalized path risks ($\pathRisk$) are defined as:
\begin{equation}
    \travPathDeviation = \frac{\travLength - \pathLength}{\travLength}, \quad \pathRisk = \frac{1}{\pathLength}\sum_{i=1}^{n-1}\varWeight{i+1,}{i},
\end{equation}
where $\travLength$ is the actual traveled distance.
$\travPathDeviation$ represents how similar the actual traveled distance is to the planned path length.
A larger $\travPathDeviation$ indicates frequent staggering, slipping, or struggling during travel.
To avoid overestimating $\travPathDeviation$, we set $\travPathDeviation = 0$ when it is negative.
$\pathRisk$ is derived from the weights of TRG edges along the planned path, to reflect the safety level of the path.
\section{Results and Discussions}\label{sec:results}
The results of the experiments support our claims that TRG-planner (\lowerRoman{1}) successfully configures environments as an efficient graph structure, (\lowerRoman{2}) finds a safer path considering the relative risk and reachability faster than existing methods, and (\lowerRoman{3}) operates effectively in a real quadrupedal robot.

\begin{table}[t!]
    \caption{
        Comparison of the proposed method according to safety factor~$\safetyGain$.
        The unit of $\pathLength$ and $\srt$ is $[\meterUnit]$ and $[\percentUnit]$, respectively.
        Other metrics are dimensionless. The \tbf{bold} values indicate the best performance for each metric.
        }
    \label{tab:risk_comparison}
    \centering
    \captionsetup{font=footnotesize}
    \begin{tabular}{c|l@{\hskip 3pt}l|c|c|c|c}
        \toprule
        \multicolumn{3}{c|}{Metric} & $\pathLength$ ($\downarrow$) & $\pathRisk$ ($\downarrow$) & $\travPathDeviation$ ($\downarrow$) & $\srt$ ($\uparrow$) \\
        \midrule[\heavyrulewidth]
        \multirow{3}{*}{\rotatebox{90}{\texttt{Short}}}  & Opti. &($\safetyGain=1.0$)     & \tbf{10.43} & 0.235       & 0.137       & \tbf{79.8}    \\ 
                                                & Cons. &($\safetyGain=10.0$)  & 11.46       & \tbf{0.125} & 0.114       & \tbf{79.8}    \\ 
                                                & Bal. &($\safetyGain=3.0$)      & 11.07       & 0.170       & \tbf{0.112} & \tbf{79.8}    \\ 
        \midrule
        \multirow{3}{*}{\rotatebox{90}{\texttt{Medium}}} & Opti. &($\safetyGain=1.0$)    & \tbf{20.99} & 0.202       & 0.078       & 78.8          \\ 
                                                & \rule{0pt}{2ex}Cons. &($\safetyGain=10.0$)  & 22.59       & \tbf{0.104} & 0.080       & 81.8          \\ 
                                                & \rule{0pt}{2ex}Bal. &($\safetyGain=3.0$)      & 21.31       & 0.142       & \tbf{0.069} & \tbf{83.8}    \\ 
        \midrule
        \multirow{3}{*}{\rotatebox{90}{\texttt{Long}}}   & Opti. &($\safetyGain=1.0$)    & \tbf{32.15} & 0.189       & 0.056       & 68.7          \\ 
                                                & Cons. &($\safetyGain=10.0$)  & 34.84       & \tbf{0.097} & 0.052       & 70.7          \\ 
                                                & Bal. &($\safetyGain=3.0$)      & 32.71       & 0.132       & \tbf{0.048} & \tbf{79.8}    \\ 
        \bottomrule
    \end{tabular}
    \vspace{-0.2cm}
\end{table}

\subsection{Comparison of Safety Strategy}\label{sec:comparison_by_safety_strategy}
\tabref{tab:risk_comparison} shows the comparison of the proposed TRG-planner with the safety strategies according to the safety factor~$\safetyGain$.
$\pathLength$ of the optimistic strategy, which prefers shorter paths, was always the shortest among the three strategies.
However, the shortest path sometimes neglects the traversal risk, resulting in high $\travPathDeviation$ and low $\srt$.
The conservative strategy focuses on safer paths rather than shorter paths, resulting in the lowest $\pathRisk$.
Unfortunately, it sometimes excessively avoids traversal risks, yielding a lengthy path that negatively affects $\travPathDeviation$ and $\srt$.
Unlike these two strategies, the balanced strategy achieved a good balance between $\pathLength$ and $\pathRisk$.
Remarkably, $\travPathDeviation$ and $\srt$ of the balanced strategy were always the best in all scenarios, implying that it discovered safer paths for the robot to travel without falling, even though $\pathLength$ was not the shortest.

\subsection{Comparison with Existing Methods}\label{sec:comparison_with_existing_methods}
\begin{figure}[t!]
    \captionsetup{font=footnotesize}
    \centering
    \includegraphics[width=1.0\columnwidth]{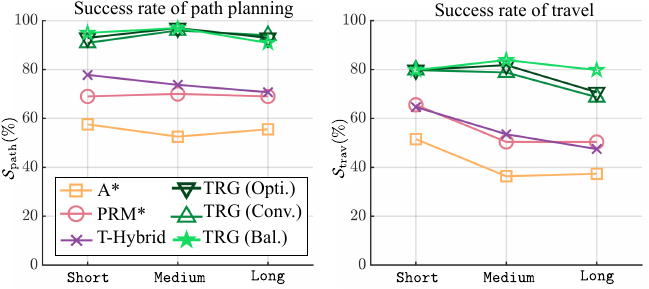}
    \caption{
        Comparison of the success rate of path planning ($\srp$) and travel ($\srt$) in the simulation environment.
        The proposed method overwhelmingly outperforms other methods in both $\srp$ and $\srt$, showing consistent $\srp$ even with increased scenario length.
        Although $\srt$ slightly reduces, this decrease is natural as the total path length increases, and the performance variation of the proposed method is smaller than other methods.
        }
    \label{fig:success_rate_graph}
\end{figure}

The $\srp$ and $\srt$ of the methods are shown in~\figref{fig:success_rate_graph}.
Existing methods struggled to plan the paths due to the highly irregular environments.
However, the proposed method, which can comprehend the entire environment by representing the direction-aware reachability of the terrain, shows a higher $\srp$ than other methods in all scenarios.
Additionally, TRG-planner achieves higher $\srt$ than other methods by accounting for the relative risk of the path during planning, which helps to minimize the risk of falling.
Furthermore, the proposed method shows consistent $\srt$ as the scenario length increases, while other methods show a decreasing tendency.

The planning time results are shown in~\tabref{tab:algorithm_time_comparison}.
Although A* could plan without bottlenecks in the \Seshort scenario, its planning time increased along with the scenario length.
PRM*, which initializes a roadmap whose quality heavily depends on the number of samples, was configured with the same number of samples as the TRG nodes.
Both PRM* and T-Hybrid methods had a competitive planning time, but experienced longer initialization times due to map preprocessing methods.
TRG-planner demonstrated the shortest planning time in all scenarios because TRG is constructed by including only essential information for safe path planning.

\begin{figure*}[t!]
    \captionsetup{font=footnotesize}
    \centering
    \includegraphics[width=1.0\textwidth]{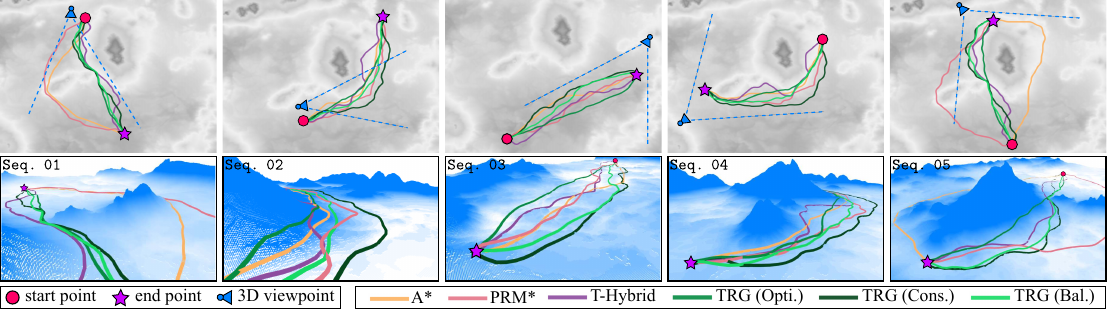}
    \caption{
        Qualitative comparison results for five sequences of the path planning simulation.
        Each planned path is represented by a different color for visualization (best viewed in color).
        The upper row shows the top view of the planned paths, showing the start (red circle) and goal (purple star) positions and a 3D viewpoint (blue camera icon).
        The lower row shows the 3D view from that camera view.
        }
    \label{fig:qualitative_result}
    \vspace{-0.2cm}
\end{figure*}

We have selected five start-goal pairs in the \Selong scenario to compare the path planning results qualitatively.
As shown in~\figref{fig:qualitative_result}, A* and PRM* sometimes excessively detoured narrow valleys because they regard such areas as collision spaces, or encroached on risky terrains to follow the shortest path.
While T-Hybrid considered a traversability through a hybrid map, it sometimes generated zigzag paths on slopes, which could be dangerous for the robot.
In contrast, TRG-planner consistently showed safe and distance-efficient paths across all scenarios by considering the relative risk of the path and the reachability of the terrain.
This is also supported by the quantitative results in~\tabref{tab:quantitative_results}, where TRG-planner, regardless of the strategy selection, ranks high in both $\pathLength$ and $\pathRisk$.

As intended, ours with the optimistic strategy shows consistently low $\pathLength$ values, which is important for path efficiency.
Additionally, ours with the conservative strategy shows low $\pathRisk$ values, which indicates high safety level.
Consequently, TRG-planner with the balanced strategy shows the best $\travPathDeviation$ in all scenarios (except for a few cases, where it is the second-best by a small margin), indicating that the robot can travel the path efficiently and safely.

\begin{table}[t!]
    \centering
    \caption{
        Comparison of the planning time according to the algorithm and scenario.
        The \tbf{bold} values indicate the best performance for each metric.
        }
    \captionsetup{font=footnotesize}
    \label{tab:algorithm_time_comparison}
    \begin{tabular}{c|l|S[table-format=4.2]S[table-format=2.2]S[table-format=3.2]S[table-format=1.2]}
    \toprule
    \multicolumn{2}{c|}{Method}          & \text{A*}     & \text{PRM*}   & \text{T-Hybrid} & \text{TRG}    \\
    \midrule[\heavyrulewidth]
    \multicolumn{2}{c|}{Initialization time $[\secUnit]$} & \text{-} & 37.86 & 30.00   & \tbf{4.01} \\
    \midrule
    \multirow{2}{*}{Planning}   & \texttt{Short}   & 350.60 & 3.50 & 86.39   & \tbf{0.43} \\
    \multirow{2}{*}{ time $[\msecUnit]$}  & \texttt{Medium}  & 1547.23 & 8.83 & 254.31   & \tbf{1.35} \\
                                & \texttt{Long}    & 2980.02 & 16.89 & 482.28   & \tbf{3.00} \\
    \bottomrule
    \end{tabular}
    \vspace{-0.2cm}
\end{table}


\begin{table*}[t!]
    \centering
    \caption{
        Quantitative evaluation on the five sequences of the simulation environment.
        Lower values of $\pathLength$, $\pathRisk$, and $\travPathDeviation$ are desirable regarding a safe and efficient navigation.
        The best results are indicated in \tbf{bold}, and the second-best results are \uli{underlined}.
        }
    \captionsetup{font=footnotesize}
    \label{tab:quantitative_results}
    \begin{tabular}{@{}c| *{4}{c|S[table-format=1.3]|c|} c|S[table-format=1.3]|c@{}}
    \toprule
    Sequence & \multicolumn{3}{c|}{\texttt{Seq. 01}} & \multicolumn{3}{c|}{\texttt{Seq. 02}} & \multicolumn{3}{c|}{\texttt{Seq. 03}} & \multicolumn{3}{c|}{\texttt{Seq. 04}} & \multicolumn{3}{c}{\texttt{Seq. 05}} \\
    \cmidrule(lr){2-4} \cmidrule(lr){5-7} \cmidrule(lr){8-10} \cmidrule(lr){11-13} \cmidrule(lr){14-16}
    Metrics & $\pathLength$ & \text{$\pathRisk$} & $\travPathDeviation$ & \text{$\pathLength$} & $\pathRisk$ & $\travPathDeviation$ & \text{$\pathLength$} & $\pathRisk$ & $\travPathDeviation$ & \text{$\pathLength$} & $\pathRisk$ & $\travPathDeviation$ & \text{$\pathLength$} & $\pathRisk$ & $\travPathDeviation$ \\
    \midrule[\heavyrulewidth]

    \mc{A*}           & \text{41.61}       & \text{0.271}       & \text{0.063}       & \uli{35.23} & \text{0.320}       & \text{0.135}       & \uli{33.58} & \text{0.228}       & \text{0.142}       & \tbf{35.52} & \text{0.457}       & \text{0.121}       & \text{42.85}       & \text{0.388}       & \text{0.073}       \\
    \mc{PRM*}         & \text{42.33}       & \text{0.303}       & \text{0.051}       & \text{36.69}       & \text{0.237}       & \text{0.088}       & \tbf{33.36}       & \text{0.301}       & \text{0.174}       & \text{37.04}       & \text{0.407}       & \text{0.104} & \text{44.40}       & \text{0.162}       & \text{0.063}       \\
    \mc{T-Hybrid}     & \text{35.11}       & \text{0.392}       & \text{0.047}       & \text{37.14}       & \text{0.410}       & \text{0.094}       & \text{35.13}       & \text{0.417}       & \text{0.070}       & \text{39.78}       & \text{0.431}       & \text{0.067}       & \text{38.03}       & \text{0.393}       & \tbf{0.059} \\
    \mc{TRG (Opti.)}  & \tbf{32.26} & \text{0.173}       & \uli{0.028} & \tbf{34.92} & \text{0.255}       & \text{0.062}       & \text{33.96} & \text{0.132}       & \text{0.075}       & \uli{35.80} & \text{0.309}       & \text{0.068}       & \tbf{33.87} & \text{0.388}       & \text{0.065}       \\
    \mc{TRG (Cons.)}  & \text{34.51}       & \tbf{0.111} & \text{0.038}       & \text{39.84}       & \tbf{0.085} & \uli{0.060} & \text{36.02}       & \tbf{0.108} & \tbf{0.031} & \text{42.58}       & \tbf{0.087} & \uli{0.063}       & \text{39.46}       & \tbf{0.111} & \text{0.070}       \\
    \mc{TRG (Bal.)}  & \uli{33.43} & \uli{0.134} & \tbf{0.018} & \text{35.71}       & \uli{0.172} & \tbf{0.054} & \text{34.20}       & \uli{0.123} & \uli{0.038} & \text{37.19}       & \uli{0.251} & \tbf{0.046} & \uli{37.47} & \uli{0.147} & \uli{0.061} \\
    
    \bottomrule
    \end{tabular}
    \vspace{-0.25cm}
\end{table*}

\begin{figure*}[t!]
    \captionsetup{font=footnotesize}
    \centering
    \includegraphics[width=1.0\textwidth]{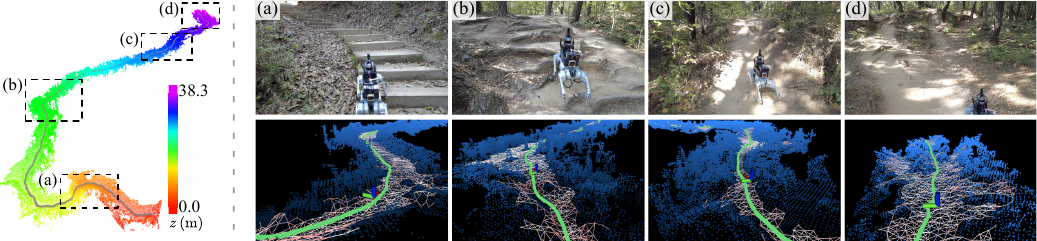}
    \caption{
        TRG-planner generated paths from a real mountainous environment.
        The first column shows the overall map of the environment and the robot's complete path.
        A safe and efficient path was successfully planned for each section: (a) a slope with steps and side paths, (b) a wide rough region with tangled tree roots, (c) a narrow rough passage, and (d) a three-way region.
        }
    \label{fig:real_mountain_result}
    \vspace{-0.2cm}
\end{figure*}
\begin{figure}[t!]
    \captionsetup{font=footnotesize}
    \centering
    \includegraphics[width=0.9\columnwidth]{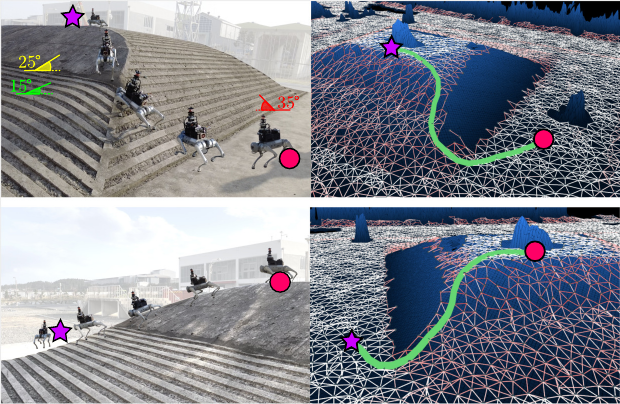}
    \caption{
        The mound has different slopes on the left ($15\degrees$ and $25\degrees$) and right ($35\degrees$) sides.
        A robot successfully found the safe entry direction using TRG-planner and climbed up (upper figure) and down (lower figure) the mound from start (red circle) to goal (purple star) positions.
        }
    \label{fig:pohang_result}
\end{figure}
\begin{figure*}[t!]
    \captionsetup{font=footnotesize}
    \centering
    \includegraphics[width=1.0\textwidth]{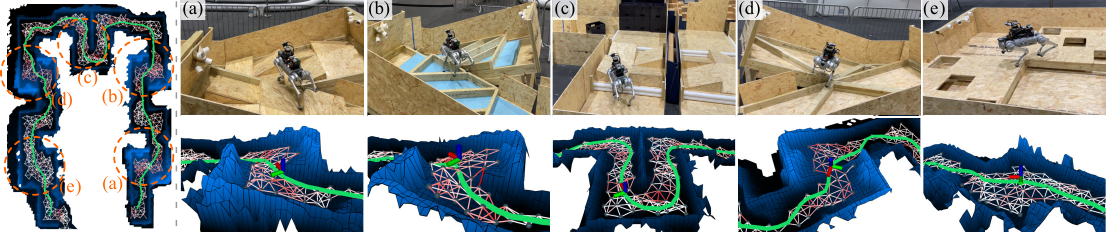}
    \caption{
        TRG-planner generated the global path from the QRC competition arena (left column) and (a-e) the detailed path planning process.
        The robot successfully traversed the difficult (a) sloped ramps, (b) soft floor with step-overs, (c) steps with pipes, (d) slope with K-rails, and (e) crate-mimic terrain.
        TRG figured out a safe and efficient path by optimizing the graph (green lines).
        }
    \label{fig:qrc_result}
    \vspace{-0.35cm}
\end{figure*}

\vspace{-0.2cm}
\subsection{Real-world Results}\label{sec:real_world_results}
In the mountainous environment, TRG-planner successfully decomposed the unstructured terrain into an efficient graph structure according to the traversal risk of the terrain, as shown in~\figref{fig:real_mountain_result}.
Although there are various irregular terrains, such as slopes, stairs, and narrow walkways, TRG-planner successfully planned a safe path, covering a total distance of $245\meterUnit$.
In~\figref{fig:pohang_result}, TRG are only connected in a gentle slope direction.
This is because TRG considered the relative risk of the path and connected the nodes, thus enabling the robot to find a safe uphill path.
The robot successfully climbed up and down the mound autonomously using the safest entry direction.

TRG-planner was also tested in the QRC arena (\figref{fig:qrc_result}) to demonstrate its ability to plan a safe path by addressing the relative traversal risk.
The QRC arena is hard to navigate due to various terrains and $15\degrees$ X-shaped slope crossroads.
Nevertheless, TRG-planner successfully planned a safe global path and autonomously navigated the robot across the arena, leading the DreamSTEP team to win this challenge.
\vspace{-0.2cm}
\section{Conclusion and Future Work}\label{sec:conclusion}
In this paper, we proposed a safe and distance-efficient path planning method called TRG-planner by capturing the reachability of the terrain and the relative risk of path candidates.
The results through simulation and real-world experiments further validated safe and efficient navigation behaviors.
The design feature of TRG-planner lies in its direction-aware risk, which is specifically tailored to non-holonomic robots and is effective in addressing such challenges.
In the future, we plan to extend the applicability of TRG-planner to accommodate a wider range of platform types.

\bibliographystyle{URL-IEEEtrans}
\vspace{-0.3cm}
\bibliography{URL-bib}

\begin{thebibliography}{10}
\providecommand{\url}[1]{#1}
\csname url@rmstyle\endcsname
\providecommand{\newblock}{\relax}
\providecommand{\bibinfo}[2]{#2}
\providecommand\BIBentrySTDinterwordspacing{\spaceskip=0pt\relax}
\providecommand\BIBentryALTinterwordstretchfactor{4}
\providecommand\BIBentryALTinterwordspacing{\spaceskip=\fontdimen2\font plus
\BIBentryALTinterwordstretchfactor\fontdimen3\font minus
  \fontdimen4\font\relax}
\providecommand\BIBforeignlanguage[2]{{%
\expandafter\ifx\csname l@#1\endcsname\relax
\typeout{** WARNING: IEEEtran.bst: No hyphenation pattern has been}%
\typeout{** loaded for the language `#1'. Using the pattern for}%
\typeout{** the default language instead.}%
\else
\language=\csname l@#1\endcsname
\fi
#2}}

\bibitem{yang2022far}
F.~Yang, C.~Cao, H.~Zhu, J.~Oh, and J.~Zhang, ``{FAR Planner: Fast, attemptable
  route planner using dynamic visibility update},'' in \emph{Proc. IEEE/RSJ
  Int. Conf. Intell. Robot. Syst.}, 2022, pp. 9--16.

\bibitem{kim2022learning}
Y.~Kim, C.~Kim, and J.~Hwangbo, ``{Learning forward dynamics model and informed
  trajectory sampler for safe quadruped navigation},'' in \emph{Robot. Sci.
  Syst.}, 2022, doi: https://doi.org/10.15607/rss.2022.xviii.069.

\bibitem{jian2021global}
Z.~Jian, S.~Zhang, S.~Chen, Z.~Nan, and N.~Zheng, ``{A global-local coupling
  two-stage path planning method for mobile robots},'' \emph{IEEE Robot.
  Automat. Lett.}, vol.~6, no.~3, pp. 5349--5356, 2021.

\bibitem{nakajima2011rt}
S.~Nakajima, ``{RT-Mover: a rough terrain mobile robot with a simple leg--wheel
  hybrid mechanism},'' \emph{Int. J. Robot. Res.}, vol.~30, no.~13, pp.
  1609--1626, 2011.

\bibitem{xu2020high}
K.~Xu, S.~Wang, X.~Wang, J.~Wang, Z.~Chen, and D.~Liu, ``{High-flexibility
  locomotion and whole-torso control for a wheel-legged robot on challenging
  terrain},'' in \emph{Proc. IEEE Int. Conf. Robot. Automat.}, 2020, pp.
  10\,372--10\,377.

\bibitem{nahrendra2023dreamwaq}
I.~M.~A. Nahrendra, B.~Yu, and H.~Myung, ``{DreamWaQ: Learning robust
  quadrupedal locomotion with implicit terrain imagination via deep
  reinforcement learning},'' in \emph{Proc. IEEE Int. Conf. Robot. Automat.},
  2023, pp. 5078--5084.

\bibitem{kareer2023vinl}
S.~Kareer, N.~Yokoyama, D.~Batra, S.~Ha, and J.~Truong, ``{ViNL: Visual
  navigation and locomotion over obstacles},'' in \emph{Proc. IEEE Int. Conf.
  Robot. Automat.}, 2023, pp. 2018--2024.

\bibitem{choi2023learning}
S.~Choi, G.~Ji, J.~Park, H.~Kim, J.~Mun, J.~H. Lee, and J.~Hwangbo, ``{Learning
  quadrupedal locomotion on deformable terrain},'' \emph{Sci. Robot.}, vol.~8,
  no.~74, p. eade2256, 2023.

\bibitem{frey2022locomotion}
J.~Frey, D.~Hoeller, S.~Khattak, and M.~Hutter, ``{Locomotion policy guided
  traversability learning using volumetric representations of complex
  environments},'' in \emph{Proc. IEEE/RSJ Int. Conf. Intell. Robot. Syst.},
  2022, pp. 5722--5729.

\bibitem{miki2022elevation}
T.~Miki, L.~Wellhausen, R.~Grandia, F.~Jenelten, T.~Homberger, and M.~Hutter,
  ``{Elevation mapping for locomotion and navigation using GPU},'' in
  \emph{Proc. IEEE/RSJ Int. Conf. Intell. Robot. Syst.}, 2022, pp. 2273--2280.

\bibitem{shan2018bayesian}
T.~Shan, J.~Wang, B.~Englot, and K.~Doherty, ``{Bayesian generalized kernel
  inference for terrain traversability mapping},'' in \emph{Conf. Robot
  Learning}, 2018, pp. 829--838.

\bibitem{elfes1989computer}
A.~Elfes, ``Using occupancy grids for mobile robot perception and navigation,''
  \emph{Computer}, vol.~22, no.~6, pp. 46--57, 1989.

\bibitem{hornung2013ar}
A.~Hornung, K.~M. Wurm, M.~Bennewitz, C.~Stachniss, and W.~Burgard, ``{OctoMap:
  An efficient probabilistic 3D mapping framework based on Octrees},''
  \emph{Auton. Robot.}, vol.~34, pp. 189--206, 2013.

\bibitem{krusi2017driving}
P.~Kr{\"u}si, P.~Furgale, M.~Bosse, and R.~Siegwart, ``{Driving on point
  clouds: Motion planning, trajectory optimization, and terrain assessment in
  generic nonplanar environments},'' \emph{J. Field Robot.}, vol.~34, no.~5,
  pp. 940--984, 2017.

\bibitem{oleynikova2017voxblox}
H.~Oleynikova, Z.~Taylor, M.~Fehr, R.~Siegwart, and J.~Nieto, ``{Voxblox:
  Incremental 3D Euclidean signed distance fields for on-board MAV planning},''
  in \emph{Proc. IEEE/RSJ Int. Conf. Intell. Robot. Syst.}, 2017, pp.
  1366--1373.

\bibitem{fankhauser2014robot}
P.~Fankhauser, M.~Bloesch, C.~Gehring, M.~Hutter, and R.~Siegwart,
  ``{Robot-centric elevation mapping with uncertainty estimates},'' in
  \emph{Mob. Serv. Robot.}, 2014, pp. 433--440.

\bibitem{fankhauser2018probabilistic}
P.~Fankhauser, M.~Bloesch, and M.~Hutter, ``{Probabilistic terrain mapping for
  mobile robots with uncertain localization},'' \emph{IEEE Robot. Automat.
  Lett.}, vol.~3, no.~4, pp. 3019--3026, 2018.

\bibitem{oh2022travel}
M.~Oh, E.~Jung, H.~Lim, W.~Song, S.~Hu, E.~M. Lee, J.~Park, J.~Kim, J.~Lee, and
  H.~Myung, ``{TRAVEL: Traversable ground and above-ground object segmentation
  using graph representation of 3D LiDAR scans},'' \emph{IEEE Robot. Automat.
  Lett.}, pp. 7255--7262, 2022.

\bibitem{brandao2020gaitmesh}
M.~Brandao, O.~B. Aladag, and I.~Havoutis, ``{GaitMesh: controller-aware
  navigation meshes for long-range legged locomotion planning in multi-layered
  environments},'' \emph{IEEE Robot. Automat. Lett.}, vol.~5, no.~2, pp.
  3596--3603, 2020.

\bibitem{wermelinger2016navigation}
M.~Wermelinger, P.~Fankhauser, R.~Diethelm, P.~Kr{\"u}si, R.~Siegwart, and
  M.~Hutter, ``{Navigation planning for legged robots in challenging
  terrain},'' in \emph{Proc. IEEE/RSJ Int. Conf. Intell. Robot. Syst.}, 2016,
  pp. 1184--1189.

\bibitem{norby2020fast}
J.~Norby and A.~M. Johnson, ``{Fast global motion planning for dynamic legged
  robots},'' in \emph{Proc. IEEE/RSJ Int. Conf. Intell. Robot. Syst.}, 2020,
  pp. 3829--3836.

\bibitem{fan2021step}
D.~D. Fan, K.~Otsu, Y.~Kubo, A.~Dixit, J.~Burdick, and A.-A. Agha-Mohammadi,
  ``{STEP: Stochastic traversability evaluation and planning for risk-aware
  off-road navigation},'' in \emph{Robot. Sci. Syst.}, 2022, doi:
  https://doi.org/10.15607/rss.2021.xvii.021.

\bibitem{chen2023smug}
C.~Chen, J.~Frey, P.~Arm, and M.~Hutter, ``{SMUG Planner: A safe multi-goal
  planner for mobile robots in challenging environments},'' \emph{IEEE Robot.
  Automat. Lett.}, vol.~8, no.~11, pp. 7170--7177, 2023.

\bibitem{liu2023hybrid}
J.~Liu, X.~Chen, J.~Xiao, S.~Lin, Z.~Zheng, and H.~Lu, ``{Hybrid map-based path
  planning for robot navigation in unstructured environments},'' in \emph{Proc.
  IEEE/RSJ Int. Conf. Intell. Robot. Syst.}, 2023, pp. 2216--2223.

\bibitem{yoo2024traversability}
S.-W. Yoo, E.-I. Son, and S.-W. Seo, ``{Traversability-aware adaptive
  optimization for path planning and control in mountainous terrain},''
  \emph{IEEE Robot. Automat. Lett.}, vol.~9, no.~6, pp. 5078--5085, 2024.

\bibitem{wellhausen2019should}
L.~Wellhausen, A.~Dosovitskiy, R.~Ranftl, K.~Walas, C.~Cadena, and M.~Hutter,
  ``{Where should I walk? predicting terrain properties from images via
  self-supervised learning},'' \emph{IEEE Robot. Automat. Lett.}, vol.~4,
  no.~2, pp. 1509--1516, 2019.

\bibitem{wellhausen2020safe}
L.~Wellhausen, R.~Ranftl, and M.~Hutter, ``{Safe robot navigation via
  multi-modal anomaly detection},'' \emph{IEEE Robot. Automat. Lett.}, vol.~5,
  no.~2, pp. 1326--1333, 2020.

\bibitem{guan2022ga}
T.~Guan, D.~Kothandaraman, R.~Chandra, A.~J. Sathyamoorthy, K.~Weerakoon, and
  D.~Manocha, ``{GA-Nav: Efficient terrain segmentation for robot navigation in
  unstructured outdoor environments},'' \emph{IEEE Robot. Automat. Lett.},
  vol.~7, no.~3, pp. 8138--8145, 2022.

\bibitem{guzzi2020path}
J.~Guzzi, R.~O. Chavez-Garcia, M.~Nava, L.~M. Gambardella, and A.~Giusti,
  ``{Path planning with local motion estimations},'' \emph{IEEE Robot. Automat.
  Lett.}, vol.~5, no.~2, pp. 2586--2593, 2020.

\bibitem{wellhausen2021rough}
L.~Wellhausen and M.~Hutter, ``{Rough terrain navigation for legged robots
  using reachability planning and template learning},'' in \emph{Proc. IEEE/RSJ
  Int. Conf. Intell. Robot. Syst.}, 2021, pp. 6914--6921.

\bibitem{wellhausen2023artplanner}
------, ``{ArtPlanner: Robust legged robot navigation in the field},'' \emph{J.
  Field Robot.}, vol.~3, no.~1, pp. 413--434, 2023.

\bibitem{yang2021real}
B.~Yang, L.~Wellhausen, T.~Miki, M.~Liu, and M.~Hutter, ``{Real-time optimal
  navigation planning using learned motion costs},'' in \emph{Proc. IEEE Int.
  Conf. Robot. Automat.}, 2021, pp. 9283--9289.

\bibitem{kavraki1996probabilistic}
L.~E. Kavraki, P.~Svestka, J.-C. Latombe, and M.~H. Overmars, ``{Probabilistic
  roadmaps for path planning in high-dimensional configuration spaces},''
  \emph{IEEE Trans. Robot.}, vol.~12, no.~4, pp. 566--580, 1996.

\bibitem{hart1968formal}
P.~E. Hart, N.~J. Nilsson, and B.~Raphael, ``{A formal basis for the heuristic
  determination of minimum cost paths},'' \emph{IEEE Trans. Syst. Sci.
  Cybern.}, vol.~4, no.~2, pp. 100--107, 1968.

\bibitem{xu2022fast}
W.~Xu, Y.~Cai, D.~He, J.~Lin, and F.~Zhang, ``{FAST-LIO2: Fast direct
  LiDAR-inertial odometry},'' \emph{IEEE Trans. Robot.}, vol.~38, no.~4, pp.
  2053--2073, 2022.

\bibitem{coulter1992implementation}
R.~C. Coulter, \emph{{Implementation of The Pure Pursuit Path Tracking
  Algorithm}}, Tech. Rep. CMU-RI-TR-92-01, Carnegie Mellon University, 1992.

\bibitem{karaman2011sampling}
S.~Karaman and E.~Frazzoli, ``{Sampling-based algorithms for optimal motion
  planning},'' \emph{Int. J. Robot. Res.}, vol.~30, no.~7, pp. 846--894, 2011.

\bibitem{box2015time}
G.~E. Box, G.~M. Jenkins, G.~C. Reinsel, and G.~M. Ljung, \emph{{Time series
  analysis: forecasting and control}}.\hskip 1em plus 0.5em minus 0.4em\relax
  John Wiley \& Sons, 2015.

\bibitem{jacoff2023taking}
A.~Jacoff, J.~Jeon, O.~Huke, D.~Kanoulas, S.~Ha, D.~Kim, and H.~Moon, ``{Taking
  the first step toward autonomous quadruped robots: The Quadruped Robot
  Challenge at ICRA 2023 in London},'' \emph{IEEE Robot. Automat. Mag.},
  vol.~30, no.~3, pp. 154--158, 2023.

\end{thebibliography}

\end{document}